\documentclass[10pt,twocolumn,letterpaper]{article}

\usepackage[pagenumbers]{cvpr} 

\usepackage{cuted} 

\usepackage{currfile} 

\usepackage{caption} 

\usepackage{amsmath,amssymb,amsthm}
\usepackage{graphicx,booktabs,multirow,colortbl}
\usepackage{tikz}
\usetikzlibrary{arrows.meta,positioning}

\makeatletter
\newcommand{\graytablerow}{%
  \rlap{\color[gray]{0.92}\vrule width\linewidth
    height\ht\@arstrutbox depth\dp\@arstrutbox}%
}
\makeatother

\newcommand{\method}{SF-Flow}
\newcommand{\R}{\mathbb{R}}
\newcommand{\E}{\mathbb{E}}
\newcommand{\norm}[1]{\left\lVert#1\right\rVert}

\usepackage{soul}
\sethlcolor{yellow}
\soulregister\cite7
\soulregister\ref7
\usepackage[textsize=scriptsize]{todonotes}

\definecolor{cvprblue}{rgb}{0.21,0.49,0.74}
\usepackage[pagebackref,breaklinks,colorlinks,allcolors=cvprblue]{hyperref}

\def\confName{CVPR}
\def\confYear{2027}

\title{SF-Flow: Spectral Flow Matching with High Structural Fidelity \\ for Hallucination-Suppressed Scientific Image Generation}

\author{%
  Zehua Wang\textsuperscript{1}\quad
  Patxi Fernandez-Zelaia\textsuperscript{2}\quad
  Mohamed Wahib\textsuperscript{3}\quad
  Xiao Wang\textsuperscript{2}\\
  Isaac Lyngaas\textsuperscript{2}\quad
  Amir Koushyar Ziabari\textsuperscript{2}\quad
  Ferdinando Fioretto\textsuperscript{1}\\
  \textsuperscript{1}University of Virginia\quad
  \textsuperscript{2}Oak Ridge National Laboratory\quad
  \textsuperscript{3}RIKEN\\
  {\small\texttt{\{wjq8vw,fioretto\}@virginia.edu}\quad
   \texttt{mohamed.attia@riken.jp}}\\
  {\small\texttt{\{fernandezzep,wangx2,lyngaasir,ziabariak\}@ornl.gov}}%
}

\begin{document}
\maketitle
\begin{abstract}
Scientific image generation focuses on reproducing experimental appearance while following a prescribed spatial layout. A key challenge is acute data sparsity, which requires operating with unpaired collections of materials (real images and prescribed layouts), where no direct supervision is feasible.
Work on unpaired translation has shown that it is possible to reproduce the texture and intensity variations of real images, but it is extremely hard to retain the intended phase boundaries in the prescribed locations.
This paper tackles this challenge and introduces \method, a flow-matching model trained only on real material images that imposes spectral constraints during generation, producing images that are high-fidelity and consistent with the prescribed layouts. Because layouts enter only at generation time, the approach extends to layout maps outside the distribution of any available training layouts.
We evaluate the method on synchrotron tomography of concrete, lens-coupled laboratory microCT of powders, and industrial XCT of additively manufactured metal.
Without instrument-specific modeling, we show that \method\ achieves substantially higher template agreement and joint fidelity--structure score than previous methods, including contrastive unpaired translation. 
\end{abstract}

\section{Introduction}
\label{sec:introduction}

Synthetic images with known structure are increasingly important in computational materials science and other scientific domains, where they provide labeled data for quantitative image analysis. Here, access to accurate labels associated to segmented images is key as they determine measurements such as porosity, defect statistics, and phase fractions. These data support training segmentation models~\cite{ziabari2022simurgh, kirillov2023sam} and enable algorithm testing and studies of how microstructure affects material properties~\cite{kench2021slicegan,lee2024diffusion}, reducing dependence on costly experiments and broadening the designs that can be examined \emph{in silico}.

However, obtaining realistic images with reliable labels remains a key challenge. Experimental scans are costly, and manual annotation is time-consuming and error prone, especially where limited resolution makes phase boundaries difficult to locate. 
In this context, physics-based simulators have been exploited to generate material layouts with known labels, but their images differ from experimental scans in texture, intensity variation, and noise~\cite{ziabari2025constrained}. Experimental images and prescribed layouts therefore form \emph{unpaired collections}. 
Thus, to generate realistic scientific images that preserve a prescribed material layout, one has to face the problem of linking a layout to its experimental appearance, without direct supervision. 

\begin{figure}[!t]
\centering
\includegraphics[width=\columnwidth]{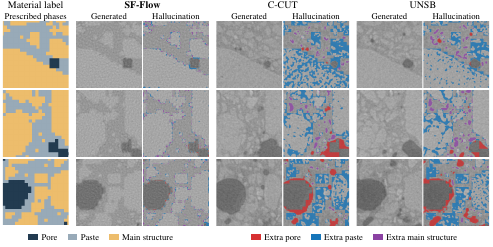}
\caption{Phase hallucinations on three concrete layouts (rows). Each material map is followed by generated images and disagreement overlays. Colors identify extra predicted pore (red), paste (blue), and main structure (purple) outside their prescribed regions.}
\label{fig:phase_mismatch}
\end{figure}

This is a fundamental problem which has been studied in several directions. First, noting that unpaired translation can give simulated images the texture and intensity variation of real scans, a cycle-consistent generative adversarial network (CycleGAN) encourages content preservation through cycle consistency~\cite{zhu2017}, and constrained contrastive unpaired translation (C-CUT) adapts patchwise contrastive learning~\cite{park2020} to concrete X-ray computed tomography (XCT)~\cite{ziabari2025constrained}. Curating simulations to match experimental feature sizes and densities~\cite{tabassum2024sam,ziabari2025constrained} requires knowing the statistics that segmentation is meant to measure. When these statistics differ, matching the target distribution can favor changes to the layout. Cohen et al.~\cite{cohen2018distribution} found that changing tumor prevalence in the target set could lead a translator to add or remove tumor-like features in synthetic magnetic resonance (MR) images of the brain. In synchrotron concrete, the same objective can favor larger pores if they are more common in the experimental collection. In that regime, a pore can look realistic while extending beyond its prescribed boundary, as the red overlays in Fig.~\ref{fig:phase_mismatch} illustrate.


However, while these methods preserve image realism they do not capture the prescribed layout precisely. Distributional scores such as FID do not establish image-map correspondence, and structural guidance applied during sampling can degrade texture and intensity statistics. Yet the exact structural layout is essential: material properties such as porosity, connectivity, and transport depend on the precise location of phases and boundaries, so a realistic image with the wrong layout is of little scientific value.

We address this gap with \method, which trains a flow-matching model~\cite{lipman2023,liu2023} on a single data modality, real images, and imposes layout constraints only at generation time. The pretrained model serves as an appearance prior and requires no paired data. Our key technical idea is a spectral constraint in Fourier space: we compare each generated image with an intensity template derived from its material map and bound their difference within a selected low-frequency band. This controls coarse spatial agreement and leaves higher-frequency texture and noise unconstrained. The constraint is enforced by viewing sampling as an optimization process, where, at selected steps, a finite primal-dual solver finds the smallest change to the predicted clean image that satisfies the constraints, and the sampler then resumes integration from the corrected state. To retain fidelity, we add a cohort-level constraint that bounds the maximum mean discrepancy~\cite{gretton2012} between generated and real Inception features, so each structural correction accounts for its effect on the feature statistics of the generated set.


We demonstrate the ability of \method\ to generate realistic images that adhere closely to prescribed layouts, on real examples of synchrotron concrete, lens-coupled laboratory microCT of powder, and industrial XCT of additively manufactured metal. 
In each case, \method\ produces higher Multi-Scale Structure Similarity and lower phase hallucination rates for each key segmentation region of the image that are important for downstream use. 

\noindent \textbf{Contributions.} We make three contributions:
{\bf (1)} We introduce SF-Flow, a framework for scientific image synthesis from unpaired material maps and experimental scans that trains an unconditional flow learns from unlabeled scans and imposes layout constraints at generation time.
{\bf (2)} We couple a spectral constraint on each image's layout with an MMD constraint on cohort fidelity, using a primal--dual solver to approximate the joint projection during generation and at the final state.
{\bf (3)} We evaluate image fidelity and layout agreement on synchrotron XCT of concrete, lens-coupled laboratory microCT of powder, and industrial XCT of additively manufactured metal. On concrete, we also evaluate higher-resolution synthesis and SAM adaptation.
\section{Related Work}
\label{sec:related}

\noindent \textbf{Synthetic Labeled Data for Scientific Imaging.}
Researchers train segmentation and reconstruction networks on synthetic image--label pairs when experimental ground truth is unavailable. In XCT of additively manufactured parts, physics-based simulators render CAD models with embedded defect libraries for artifact suppression~\cite{ziabari2022simurgh,ziabari2020aict}, and GAN-translated simulations support SAM adaptation for multiclass flaw segmentation~\cite{tabassum2024sam}. Simulated microstructures with degraded reconstructions likewise support foundation-model adaptation on gigapixel tomography~\cite{zhang2026sgma}. Researchers use comparable pipelines in fluorescence microscopy~\cite{fu2018three} and randomized-contrast label-map synthesis in medical imaging~\cite{billot2023synthseg}. These approaches supply labels, but retain simulation's appearance gap or translate between collections with similar content statistics.

\noindent \textbf{Unpaired Translation and Scientific Synthesis.}
Unpaired translators address this appearance gap by inferring correspondence from source images, whereas semantic synthesis uses layouts to guide object placement and shape in natural images~\cite{park2019}. CycleGAN uses cycle consistency~\cite{zhu2017} and CUT matches corresponding patch features~\cite{park2020}; other objectives preserve geometry under spatial transformations~\cite{fu2019gcgan} or semantic consistency across domains with differing content~\cite{jia2021srunit}. Yet Cohen et al.~\cite{cohen2018distribution} found that overall anatomy can survive translation while individual features appear or disappear. Material generators face this distinction: adversarial models reproduce pore statistics~\cite{goodfellow2014,mosser2017reconstruction} and design microstructures~\cite{yang2018microstructure}, while diffusion models reconstruct morphologies~\cite{lee2024diffusion}, but matching statistics does not establish image--map correspondence. C-CUT adds a label channel and a label-consistency loss to encourage microstructure preservation~\cite{ziabari2025constrained}. Even so, a pore can match its prescribed area while extending into paste, giving a segmenter incorrect paste labels for pore pixels. FID~\cite{heusel2017} compares collections and cannot detect this per-pair error.

\begin{figure*}[t]
\centering
\includegraphics[width=0.92\textwidth]{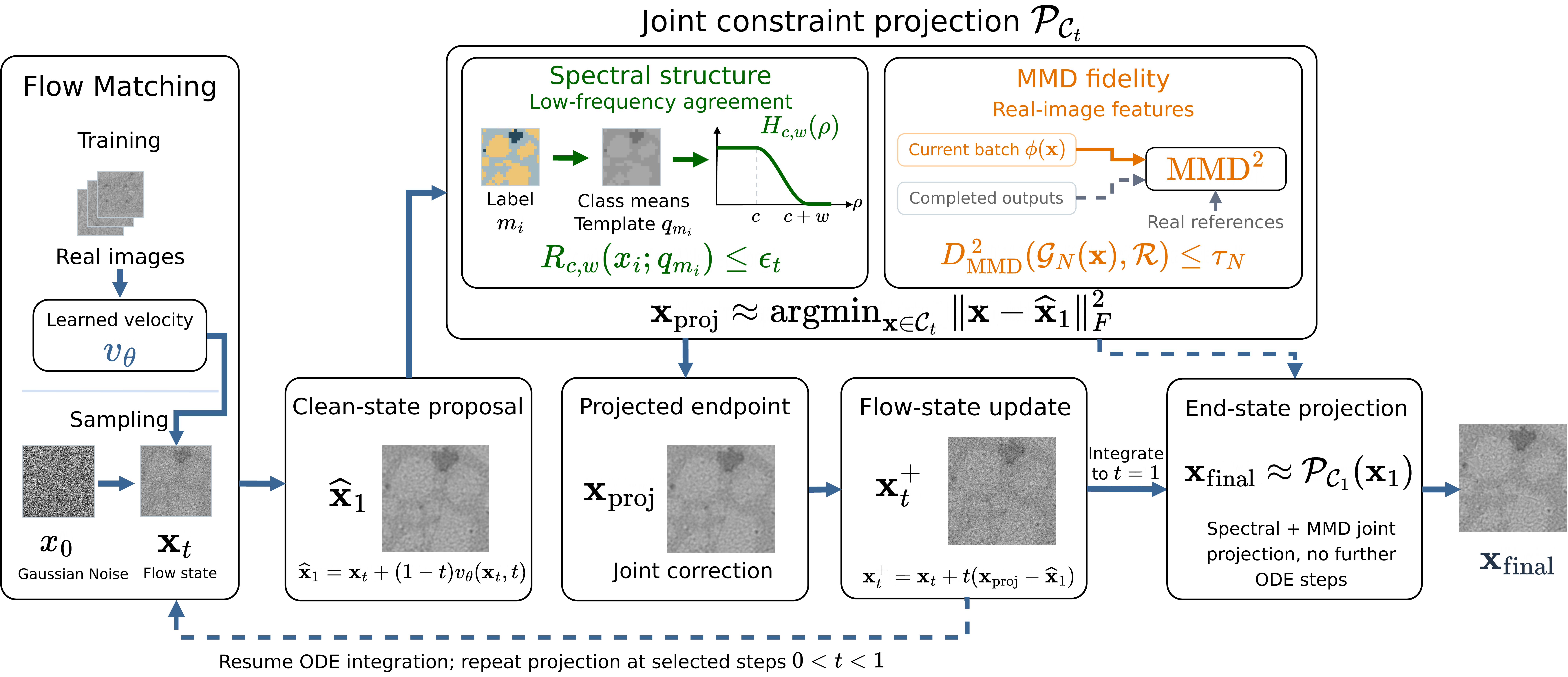}
\caption{Overview of \method. A real-image flow prior predicts clean endpoints during sampling. Joint projection seeks the smallest correction satisfying per-image low-frequency template agreement (green) and cohort MMD fidelity (orange), using active predictions, completed outputs, and real references. Corrections update the flow state before integration resumes. A final joint projection at $t=1$ precedes encoding, with no further ODE steps.}
\label{fig:overview}
\end{figure*}

\noindent \textbf{Flow Matching and Constrained Sampling.}
Researchers can instead impose requirements while sampling from diffusion~\cite{ho2020ddpm,song2021sde,rombach2022ldm} or flow-matching~\cite{lipman2023,liu2023} priors. In inverse problems, observations guide sampling trajectories~\cite{chung2023dps,zhang2024ictm} or constrain known pixels and measured components~\cite{lugmayr2022repaint,kawar2022ddrm,wang2023ddnm}; other methods optimize initial noise~\cite{benhamu2024dflow} or incorporate projection into training~\cite{christopher2026cafm}. Sampling-time methods evaluate constraints on decoded images~\cite{zampini2025trainingfree} or express them as chance constraints on intermediate states~\cite{liang2025ccfm}. Projected Diffusion Models (PDM) project the current sample after each Langevin update~\cite{christopher2024pdm}, whereas Physics-Constrained Flow Matching (PCFM) corrects a predicted clean endpoint and transfers the correction to the current flow state~\cite{utkarsh2026pcfm}. Structural correction alone need not match the texture and noise of real XCT cohorts.

We address this distinction by pairing per-image spectral guidance with cohort-level MMD during sampling. Earlier generative models use MMD for distribution matching during training~\cite{gretton2012,li2015gmmn}, and Kernel Inception Distance uses it for evaluation~\cite{binkowski2018}. In \method, the spectral bound limits low-frequency differences from each map template, while a shared MMD bound compares real features with those of the active batch and completed outputs.


\section{Problem Setting and Preliminaries}
\label{sec:setting}

\subsection{Unpaired Datasets}
We consider XCT generation from separate collections of material maps and experimental images. Let $\Omega$ be a pixel grid with $n=|\Omega|$. A phase map $m:\Omega\rightarrow\{1,\ldots,K\}$ assigns one material class to each pixel. Our main case study, synchrotron tomography of concrete, uses $K=3$ classes: pore, paste, and main structure. The map collection $\mathcal D_{\rm geom}=\{m_i\}_{i=1}^{N_g}$ specifies the requested material layouts. The image collection $\mathcal D_{\rm xct}=\{r_j\}_{j=1}^{N_{\rm xct}}$ contains unannotated images from the experimental distribution $p_{\rm xct}$. The collections are unpaired: no real image $r_j$ is assumed to match a particular map $m_i$. Nor do we assume that the maps share the feature-size, density, or spatial statistics of the real images: simulated microstructures and real specimens may differ in content as well as appearance.

We scale XCT intensities to $[-1,1]$ before training and sampling; $r_j$ and $x$ denote images in this model-space range. A simulated grayscale image accompanies each material map and is reserved for evaluation of \method. The velocity model learns from real images alone. Material maps enter through sampling-time constraints, while the trained model remains fixed.

The goal is a pair in which the image is indistinguishable from experimental data and the map is a correct segmentation of the image. Writing $S$ for an ideal segmenter, we seek generated cohorts that follow the experimental distribution and satisfy
\begin{equation}
 S(x_i)=m_i\quad\text{for every generated image } x_i.
 \label{eq:goal}
\end{equation}
Neither requirement can be evaluated directly. We replace the distributional requirement with a cohort-level feature constraint and Eq.~\eqref{eq:goal} with a per-image spectral constraint against an intensity template derived from each map (Sec.~\ref{sec:method}). The spectral constraint controls agreement only above a spatial scale set by its cutoff, so we also measure label agreement directly (Sec.~\ref{sec:experiments}).

\subsection{Flow Matching for Material Generation}
We train an unconditional flow model on real XCT images. For each example, we independently draw Gaussian noise $x_0\sim\mathcal N(0,I)$, a real image $x_1\sim p_{\rm xct}$, and a time $t\sim\mathcal U[0,1]$. A U-Net with time-conditioned residual blocks and encoder--decoder skip connections predicts the velocity $v_\theta(x_t,t)$ along the straight path from $x_0$ to $x_1$. We fit the velocity field with the flow-matching loss
{\small
\begin{equation}
 x_t=(1-t)x_0+tx_1,\;
 \mathcal L_{\rm FM}=\E\norm{v_\theta(x_t,t)-(x_1-x_0)}_2^2.
 \label{eq:fm}
\end{equation}
}

At sampling time, we draw $x_0$ from the same Gaussian source and integrate $\frac{d x_t}{d t}=v_\theta(x_t,t)$ from $t=0$ to $t=1$. \method\ keeps the learned velocity field fixed and guides the trajectory toward the prescribed layout by projecting predicted endpoints at selected steps. A final projection acts on the integrated image (Sec.~\ref{sec:method}).

\section{Methodology}
\label{sec:method}
An unconditional flow provides experimental appearance, but does not associate samples with a material map. A structural correction can establish this correspondence while changing texture and noise. \method\ addresses both in one projection: a spectral bound controls layout agreement over selected frequencies, and an MMD bound limits discrepancy between generated and real image features.

The sampler evaluates these constraints on predicted clean images (Fig.~\ref{fig:overview}). At selected steps, it approximates their minimum-displacement projection and transfers the correction to the current flow state. Integration then resumes from the corrected state. A final joint projection at $t=1$ acts on the integrated images before output encoding.

\subsection{Projected Flow Matching}
\label{sec:projectedflow}
The constraints describe clean images, so we apply them to an endpoint estimate derived from the current state and velocity. The straight-path formulation in Eq.~\eqref{eq:fm} gives the local estimates
\begin{equation}
 \widehat x_0=x_t-tv_\theta(x_t,t),\quad
 \widehat x_1=x_t+(1-t)v_\theta(x_t,t).
 \label{eq:endpoints}
\end{equation}
The pair $(\widehat x_0,\widehat x_1)$ represents $x_t$ on a local straight line. We assess structure and fidelity at the predicted endpoint $\widehat x_1$.

For a batch of $B$ images, write the endpoint predictions as $\widehat{\mathbf x}_1=(\widehat x_{1,1},\ldots,\widehat x_{1,B})$. Let $\mathcal C_t\subseteq\mathcal X_B$, where $\mathcal X_B=[-1,1]^{B\times n}$, be the set satisfying the material-map and cohort requirements at time $t$. The projection seeks the smallest change that places the batch in this set:
\begin{equation}
 \mathbf x_{\rm proj}
 =\underset{\mathbf x\in\mathcal C_t}{\operatorname{argmin}}
 \norm{\mathbf x-\widehat{\mathbf x}_1}_F^2.
 \label{eq:generalprojection}
\end{equation}
The Frobenius norm $\norm{\mathbf x-\widehat{\mathbf x}_1}_F^2=\sum_{i=1}^{B}\norm{x_i-\widehat x_{1,i}}_2^2$ sums squared intensity changes over the batch's pixels.

We transfer this correction to the trajectory by retaining the source estimates $\widehat{\mathbf x}_0$ and replacing $\widehat{\mathbf x}_1$ with $\mathbf x_{\rm proj}$:
\begin{equation}
 \mathbf x_t^+=(1-t)\widehat{\mathbf x}_0+t\mathbf x_{\rm proj}
      =\mathbf x_t+t(\mathbf x_{\rm proj}-\widehat{\mathbf x}_1).
 \label{eq:rebuild}
\end{equation}
The weight $t$ scales the endpoint displacement at the current flow time. We reevaluate the velocity at $\mathbf x_t^+$ and continue Heun integration from this corrected state.

Even when the corrected endpoint satisfies both constraints, the remaining Heun steps can produce a final batch that violates them. We apply the joint projection once more at $t=1$, when the endpoint prediction equals the integrated batch $\mathbf x_1$. The final spectral budget and the cohort MMD requirement define the feasible set $\mathcal C_1$. We seek the closest batch in this set, $\mathbf x_{\rm final}=\operatorname{argmin}_{\mathbf x\in\mathcal C_1}\norm{\mathbf x-\mathbf x_1}_F^2$. The update in Eq.~\eqref{eq:rebuild} then gives $\mathbf x_1^+=\mathbf x_{\rm final}$. We use this corrected batch as the continuous output without another ODE step.

\subsection{Spectral Structure Constraint}
\label{sec:spectral}

A phase map specifies material classes, while the flow generates intensities. We connect these representations through an intensity template. Let $\mu_k\in[0,1]$ be the mean experimental intensity of class $k$. The template for map $m$ assigns this mean to each region in the model's $[-1,1]$ range:
\begin{equation}
 q_m(p)=2\mu_k-1\quad\text{if }m(p)=k.
 \label{eq:structuralreference}
\end{equation}
The piecewise-constant template $q_m$ describes the requested layout without texture or imaging noise. Pixelwise matching would suppress these variations, so we compare the image and template within a selected frequency band.

Let $\mathcal F$ denote the unitary discrete Fourier transform, and let $\rho(\omega)$ be the radial frequency at coordinate $\omega$, measured in cycles per pixel. A cosine low-pass filter selects the band with cutoff $c$ and transition width $w$:
\begin{equation}
 H_{c,w}(\rho)=\begin{cases}
 1,&\rho\le c,\\
 \tfrac12(1+\cos(\pi(\rho-c)/w)),&c<\rho<c+w,\\
 0,&\rho\ge c+w.
 \end{cases}
 \label{eq:filter}
\end{equation}
For a candidate endpoint $z\in[-1,1]^n$, the structural residual relative to $q_m$ is
\begin{equation}
 R_{c,w}(z;q_m)=\frac{1}{\sqrt n}
 \left\|H_{c,w}\odot\left[\mathcal F(z)-\mathcal F(q_m)\right]\right\|_2.
 \label{eq:spectralresidual}
\end{equation}
The residual $R_{c,w}$ is the root-mean-square filtered difference. Frequencies below $c$ receive full weight; the weights taper between $c$ and $c+w$ and vanish above $c+w$. We require each active image to satisfy
\begin{equation}
 R_{c,w}(z_i;q_{m_i})\le\epsilon_t\quad\text{for every active image }i,
 \label{eq:structureconstraint}
\end{equation}
The budget $\epsilon_t$ decreases linearly across intermediate projection events, tightening structural agreement as sampling progresses. The terminal projection uses $\epsilon_1=\epsilon_{\rm final}$. Frequencies above $c+w$ remain unconstrained by this bound; the fidelity constraint controls appearance through image features. A structure $L$ pixels wide concentrates its energy near $1/(2L)$ cycles per pixel, so with default $c=0.06$ and $w=0.01$ the bound controls features wider than about seven pixels. Smaller features, such as small pores and thin paste ligaments, are placed by the prior rather than the constraint; our boundary variants (Sec.~\ref{sec:concrete_experiment}) target this regime.

\subsection{MMD Fidelity Constraint}
\label{sec:mmd_fidelity}
MMD compares the generated cohort with real reference images in the feature space of a frozen Inception network, $\phi(x)\in\R^d$. These features can reflect pore density as well as texture. If the prescribed maps contain more pores than the real images, reducing MMD can favor a lower pore density in the generated cohort. The per-image spectral bound limits how far each candidate can depart from its map at the selected frequencies. A cohort that meets those structural bounds can therefore still exceed its MMD budget.

We select a fixed reference set $\mathcal R\subseteq\mathcal D_{\rm xct}$ containing $N_r$ real images and index them within this set as $\mathcal R=\{r_j\}_{j=1}^{N_r}$. Let $\mathcal G$ denote a generated cohort. We compare their features with the cubic kernel $k(a,b)=(1+a^\top b/d)^3$. The biased empirical MMD estimator is
\begin{align}
 D_{\rm MMD}^2(\mathcal G,\mathcal R)
 &=\frac{1}{|\mathcal G|^2}\sum_{g,g'\in\mathcal G}
 k(\phi(g),\phi(g'))\nonumber\\
 &\quad+\frac{1}{N_r^2}\sum_{j,j'}k(\phi(r_j),\phi(r_{j'}))\nonumber\\
 &\quad-\frac{2}{|\mathcal G|N_r}\sum_{g\in\mathcal G}\sum_j
 k(\phi(g),\phi(r_j)).
 \label{eq:mmd}
\end{align}
The within-set sums retain diagonal self-similarity terms. This biased estimator is nonnegative and defined for a one-image prefix. The unbiased KID estimator instead omits these terms, requires at least two samples per set, and can be negative for finite samples~\cite{gretton2012,binkowski2018}. We use the biased form to apply one nonnegative statistic throughout generation.

During sampling, the current batch $\mathbf x=(x_1,\ldots,x_B)$ extends a growing prefix of generated images. The prefix has size $N$: it contains the $B$ candidates and $N-B$ completed outputs, denoted $\mathcal X_{N-B}=\{\bar x_\ell\}_{\ell=1}^{N-B}$. Together they form $\mathcal G_N(\mathbf x)=\mathcal X_{N-B}\cup\{x_i\}_{i=1}^{B}$. We cache the features of completed images and detach them from the gradient graph. Only the current batch changes under the shared constraint
\begin{equation}
 D_{\rm MMD}^2(\mathcal G_N(\mathbf x),\mathcal R)\le\tau_N.
 \label{eq:mmdconstraint}
\end{equation}
MMD depends on prefix size, so we calibrate a threshold $\tau_N$ for each $N$. We generate 500 unconditional images with the same checkpoint and batch size, permute their features 25 times, and compute biased MMD at every prefix. We set $\tau_N$ to $\gamma$ times the 75th percentile at size $N$.

\subsection{Joint Projection Solver}
The spectral bound applies to each active image, while the MMD bound couples the batch to the full prefix. Their intersection defines $\mathcal C_t$. Substituting these requirements into Eq.~\eqref{eq:generalprojection} gives
\begin{equation}
\begin{aligned}
 \mathbf x_{\rm proj}
 =\underset{\mathbf x\in\mathcal X_B}{\operatorname{argmin}}\quad&
 \frac{1}{2\sigma_d^2 Bn}\norm{\mathbf x-\widehat{\mathbf x}_1}_F^2\\
 \text{subject to}\quad& R_{c,w}(x_i;q_{m_i})\le\epsilon_t,\quad i=1,\ldots,B,\\
 &D_{\rm MMD}^2(\mathcal G_N(\mathbf x),\mathcal R)\le\tau_N.
\end{aligned}
\label{eq:jointprojection}
\end{equation}
The factor $Bn$ normalizes the displacement in Eq.~\eqref{eq:generalprojection} to a per-pixel average. We use $\sigma_d=0.1$ to set its characteristic scale in model space. This positive rescaling preserves the exact minimizer but affects the updates of a finite solver.

We approximate Eq.~\eqref{eq:jointprojection} with alternating primal and dual updates. After budget normalization, Adam updates the image batch to decrease the Lagrangian. Projected gradient ascent updates the nonnegative constraint multipliers. The cached features remain fixed during updates.

A verifier tests both constraints on every candidate with tolerance $\eta=10^{-4}$. For positive budgets, a candidate passes when $R_{c,w}^2/\epsilon_t^2-1\le\eta$ for every image and $D_{\rm MMD}^2/\tau_N-1\le\eta$ for the cohort. The solver retains the passing candidate with the smallest displacement. If none passes, it returns the least-violating candidate and records a failure. These checks assess feasibility; the finite solver is not guaranteed to find a feasible point or a global minimum.

Intermediate solves use the predicted endpoint as the displacement reference, then return to flow integration through Eq.~\eqref{eq:rebuild}. The terminal solve instead uses the integrated batch $\mathbf x_1$ and the budget $\epsilon_{\rm final}$. The verifier checks both constraints again after image quantization.

\section{Experiments}
\label{sec:experiments}
We evaluate image fidelity and agreement with prescribed layouts across three X-ray tomography regimes. Synchrotron tomography of concrete uses a quasi-monochromatic parallel beam and a lens-coupled scintillator, so its dominant artifacts are rings and phase-contrast edges. Laboratory microCT of powders uses a polychromatic source with a lens-coupled scintillator and optical magnification, adding beam hardening and photon-limited noise. Industrial XCT of additively manufactured metal uses a high-energy polychromatic cone beam and a flat-panel detector, where scatter, beam hardening, and streaking dominate. In all three, the simulated maps and real scans are unpaired and differ in feature size and density.

\begin{table}[!t]
\centering
\caption{Synchrotron concrete synthesis on 500 $256\times256$ maps. Metrics measure cohort fidelity (FID), template agreement (MS-SSIM), and their combined score ($H_R$, Eq.~\ref{eq:hr_score}). Phase hallucination is the per-phase percentage of prescribed pixels misclassified by intensity-based segmentation. Flow Matching samples it without guidance, Spectral-only retains the per-image layout constraint, and MMD-only retains the cohort feature constraint. Both single-constraint runs use \method\ batch size $B=8$ and match the corresponding constraint setting (Sec.~\ref{sec:configuration_selection}). Diffusion Translation is our XCT adaptation of InST~\cite{zhang2023inst} style-transfer approach.}
\label{tab:latest_mmd}
\scriptsize
\setlength{\tabcolsep}{1pt}
\renewcommand{\arraystretch}{1.08}
\begin{tabular*}{\columnwidth}{@{\extracolsep{\fill}}lrrrrrr@{}}
\toprule
\multicolumn{4}{c}{} & \multicolumn{3}{c}{Phase hallucination (\%, $\downarrow$)}\\
Method & $H_R\uparrow$ & FID$\downarrow$ & MS-SSIM$\uparrow$ & Pore & Paste & Main struct.\\
\midrule
\graytablerow \textbf{\method} & 0.7842 & 14.588 & 0.8565 & 6.10 & 4.75 & 2.33\\
\graytablerow \method\ + soft boundary & \textbf{0.8334} & 19.275 & \textbf{0.9379} & 11.33 & \textbf{2.59} & \textbf{1.88}\\
\graytablerow \method\ + hard boundary & 0.5326 & 26.028 & 0.6270 & 15.95 & 14.01 & 7.24\\
Flow Matching~\cite{liu2023} & -- & \textbf{9.291} & -- & -- & -- & --\\
Spectral-only~\cite{gonzalez2018dip} & 0.5097 & 71.578 & 0.8699 & 6.21 & 4.49 & 2.20\\
MMD-only~\cite{gretton2012} & 0.5476 & 9.300 & 0.5788 & 95.05 & 60.24 & 44.72\\
\midrule
C-CUT~\cite{ziabari2025constrained} & 0.7165 & 12.314 & 0.7714 & 7.91 & 14.22 & 34.49\\
C-CUT + hard boundary & 0.0000 & 206.783 & 0.6680 & \textbf{2.15} & 84.05 & 26.30\\
C-CUT + soft boundary & 0.7099 & 12.388 & 0.7647 & 10.34 & 14.54 & 38.38\\
UNSB~\cite{kim2024unsb} & 0.6998 & 10.536 & 0.7452 & 5.76 & 17.73 & 21.88\\
Diffusion Translation~\cite{zhang2023inst} & 0.5092 & 68.354 & 0.8423 & 2.26 & 8.13 & 9.11\\
PDM~\cite{christopher2024pdm} & 0.5969 & 58.108 & 0.8992 & 5.56 & 6.08 & 2.44\\
\shortstack[l]{PCFM~\cite{utkarsh2026pcfm}} & 0.6366 & 44.216 & 0.8553 & 10.44 & 7.47 & 3.92\\
\bottomrule
\end{tabular*}
\end{table}

\begin{figure*}[!t]
\centering
\includegraphics[width=\textwidth]{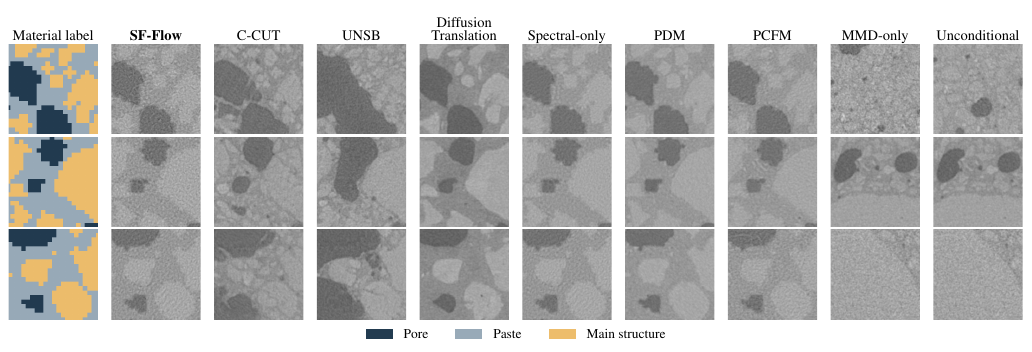}
\caption{Synchrotron concrete samples for three layouts (rows), using Table~\ref{tab:latest_mmd} settings. Map colors denote pore (dark blue), paste (gray-blue), and main structure (yellow); images share the grayscale range $[0,255]$. Spectral guidance promotes layout agreement, while MMD controls cohort appearance. MMD-only and unconditional sampling impose no per-image layout constraint.}
\label{fig:spectral_effect}
\end{figure*}

\begin{table*}[!t]
\centering
\caption{Higher-resolution concrete synthesis on 500 maps per resolution (499 readable C-CUT pairs at $1024\times1024$). Metrics measure cohort fidelity (FID), template agreement (MS-SSIM), their combined score ($H_R$), and per-phase misclassification. Flow Matching is unguided, Spectral-only constrains per-image layout, and MMD-only constrains the cohort's feature distribution. Soft-boundary \method\ reduces paste and main-structure errors but increases pore errors. At $1024\times1024$, \method\ and C-CUT use different FID references (Sec.~\ref{sec:highres_protocol}).}
\label{tab:higher_resolution}
\footnotesize
\setlength{\tabcolsep}{1pt}
\begin{tabular*}{\linewidth}{@{\extracolsep{\fill}}l*{6}{r}@{\hskip 7pt}*{6}{r}@{}}
\toprule
& \multicolumn{6}{c}{$512\times512$} & \multicolumn{6}{c}{$1024\times1024$}\\
\cmidrule(lr){2-7}\cmidrule(lr){8-13}
& \multicolumn{3}{c}{} & \multicolumn{3}{c}{Phase hallucination (\%, $\downarrow$)} & \multicolumn{3}{c}{} & \multicolumn{3}{c}{Phase hallucination (\%, $\downarrow$)}\\
Method & $H_R\uparrow$ & FID$\downarrow$ & MS-SSIM$\uparrow$ & Pore & Paste & Main struct. & $H_R\uparrow$ & FID$\downarrow$ & MS-SSIM$\uparrow$ & Pore & Paste & Main struct.\\
\midrule
\graytablerow \textbf{\method} & 0.6421 & 35.430 & 0.8077 & \textbf{8.74} & 6.88 & 3.81 & 0.3591 & 89.835 & 0.7476 & 7.30 & 6.29 & 5.52\\
\graytablerow \method\ + soft boundary & \textbf{0.7374} & 37.822 & \textbf{0.9439} & 12.79 & \textbf{2.19} & \textbf{1.91} & \textbf{0.4644} & 87.703 & \textbf{0.9427} & 10.61 & \textbf{2.04} & \textbf{2.09}\\
\graytablerow \method\ + hard boundary & 0.4383 & 52.917 & 0.6316 & 14.01 & 13.78 & 6.29 & 0.2632 & 101.244 & 0.6354 & 11.62 & 13.15 & 5.85\\
Flow Matching~\cite{liu2023} & -- & \textbf{23.876} & -- & -- & -- & -- & -- & \textbf{62.610} & -- & -- & -- & --\\
Spectral-only~\cite{gonzalez2018dip} & 0.2832 & 112.423 & 0.8101 & 8.81 & 6.86 & 3.75 & 0.0000 & 188.596 & 0.7462 & \textbf{7.13} & 6.36 & 5.61\\
MMD-only~\cite{gretton2012} & 0.4871 & 25.950 & 0.5732 & 96.09 & 58.39 & 47.57 & 0.3592 & 62.861 & 0.5645 & 96.39 & 52.91 & 50.79\\
\midrule
C-CUT~\cite{ziabari2025constrained} & 0.6240 & 26.309 & 0.7360 & 26.60 & 18.82 & 27.03 & 0.3451 & 93.479 & 0.7514 & 21.92 & 15.42 & 38.27\\
C-CUT + hard boundary & 0.0000 & 287.907 & 0.6101 & 35.18 & 81.24 & 12.57 & 0.0000 & 304.706 & 0.7647 & 22.52 & 85.75 & 49.63\\
C-CUT + soft boundary & 0.6018 & 28.653 & 0.7214 & 21.63 & 18.14 & 30.70 & 0.3446 & 92.794 & 0.7439 & 13.60 & 13.08 & 53.96\\
\bottomrule
\end{tabular*}
\end{table*}

The concrete training set contains 72,230 real crops at $256\times256$, 16,051 at $512\times512$, and 3,511 at $1024\times1024$. At $256\times256$, the metal and powder models train on 13,859 and 7,748 real crops, respectively. We reserve 5\% of the real crops in each dataset for validation. At each concrete resolution and modality, we evaluate generation on 500 prescribed maps. We also test whether generated concrete images paired with these labels support SAM adaptation.

\begin{figure}[!t]
\centering
\includegraphics[width=0.9\linewidth,trim=8 0 0 0,clip]{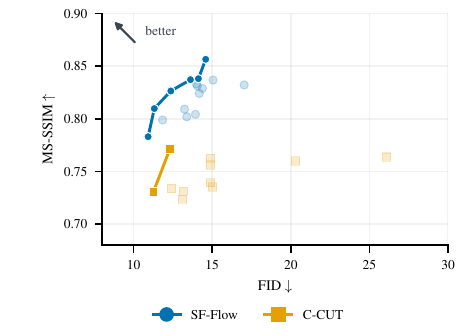}
\caption{Fidelity--structure tradeoffs on $256\times256$ concrete. Better settings have lower FID and higher template MS-SSIM (upper left). Lines connect each method's nondominated settings; faint points show other configurations. Frontiers use all points before cropping at FID 30. Search budgets differ and flow uses one seed.}
\label{fig:fid_msssim_pareto}
\end{figure}

\subsection{Synchrotron Concrete}
\label{sec:concrete_experiment}
\noindent\textbf{Fidelity and structure metrics.}
We measure cohort fidelity with Fr\'echet Inception Distance (FID)~\cite{heusel2017} on ImageNet Inception-v3 features and per-image structural agreement with multiscale structural similarity (MS-SSIM)~\cite{wang2003msssim} against the prescribed map's unsmoothed class-mean template. Because each phase has one template intensity, suppressed texture can raise MS-SSIM without improving boundaries. We combine the metrics in the normalized hypervolume score
\begin{equation}
\label{eq:hr_score}
H_R=\left(1-\frac{\mathrm{FID}}{R}\right)_+\times\mathrm{MS\text{-}SSIM},
\end{equation}
where $(a)_+=\max\{a,0\}$ and $R=172.862$ is the FID between real and simulated concrete crops at $256\times256$. We report both components because rankings depend on $R$.

\noindent\textbf{Phase hallucination.}
We assess phase placement by smoothing each saved concrete image with a Gaussian filter ($\sigma=2$ pixels) and assigning pixels to the phase with the closest calibrated mean intensity. For phase $k$, we report $(E_k/N_k)\times100\%$: $N_k$ counts prescribed phase-$k$ pixels, and $E_k$ counts those assigned another phase. A pore predicted in paste counts toward the paste rate. A prescribed pore classified as paste counts toward the pore rate. We include all region interiors and boundaries (Tables~\ref{tab:latest_mmd} and~\ref{tab:higher_resolution}).


\begin{table*}[!t]
\centering
\caption{Synchrotron concrete, industrial XCT metal AM, and laboratory microCT powder synthesis on 500 prescribed $256\times256$ maps per dataset. Flow Matching is the unguided \method\ ablation. \method\ achieves higher template MS-SSIM and joint score $H_R$ than all C-CUT variants, while standard or soft-boundary C-CUT retains lower FID. $H_R$ uses the fixed reference $R=172.862$ and is compared within datasets.}
\label{tab:additional_modalities}
\footnotesize
\setlength{\tabcolsep}{2pt}
\renewcommand{\arraystretch}{1.12}
\begin{tabular*}{\textwidth}{@{\extracolsep{\fill}}l*{9}{r}@{}}
\toprule
& \multicolumn{3}{c}{XCT Concrete} & \multicolumn{3}{c}{XCT Metal AM} & \multicolumn{3}{c}{MicroCT Powder}\\
\cmidrule(lr){2-4}\cmidrule(lr){5-7}\cmidrule(lr){8-10}
Method & $H_R\uparrow$ & FID$\downarrow$ & MS-SSIM$\uparrow$ & $H_R\uparrow$ & FID$\downarrow$ & MS-SSIM$\uparrow$ & $H_R\uparrow$ & FID$\downarrow$ & MS-SSIM$\uparrow$\\
\midrule
\graytablerow \textbf{\method} & 0.7842 & 14.588 & 0.8565 & 0.8230 & 16.860 & 0.9120 & 0.8600 & 10.807 & 0.9174\\
\graytablerow \method\ + soft boundary & \textbf{0.8334} & 19.275 & \textbf{0.9379} & \textbf{0.8276} & 17.162 & 0.9189 & \textbf{0.8729} & 11.596 & 0.9356\\
\graytablerow \method\ + hard boundary & 0.5326 & 26.028 & 0.6270 & 0.7119 & 21.635 & 0.8138 & 0.8526 & 11.263 & 0.9120\\
Flow Matching~\cite{liu2023} & -- & \textbf{9.291} & -- & -- & 9.957 & -- & -- & 15.421 & --\\
\midrule
C-CUT~\cite{ziabari2025constrained} & 0.7165 & 12.314 & 0.7714 & 0.7597 & 11.289 & 0.8127 & 0.8409 & 9.770 & 0.8913\\
C-CUT + hard boundary & 0.0000 & 206.783 & 0.6680 & 0.0000 & 292.506 & 0.6617 & 0.0000 & 243.811 & 0.8619\\
C-CUT + soft boundary & 0.7099 & 12.388 & 0.7647 & 0.7733 & \textbf{9.829} & 0.8200 & 0.8411 & 10.077 & 0.8932\\
UNSB~\cite{kim2024unsb} & 0.6998 & 10.536 & 0.7452 & 0.8024 & 11.817 & 0.8613 & 0.8402 & \textbf{9.595} & 0.8896\\
Diffusion Translation~\cite{zhang2023inst} & 0.5092 & 68.354 & 0.8423 & 0.7256 & 35.301 & 0.9118 & 0.4741 & 84.550 & 0.9279\\
PDM~\cite{christopher2024pdm} & 0.5969 & 58.108 & 0.8992 & 0.7131 & 45.029 & 0.9643 & 0.7425 & 36.082 & 0.9384\\
\shortstack[l]{PCFM~\cite{utkarsh2026pcfm}} & 0.6366 & 44.216 & 0.8553 & 0.7059 & 46.698 & \textbf{0.9671} & 0.5834 & 66.025 & \textbf{0.9439}\\
\bottomrule
\end{tabular*}
\end{table*}

\noindent{\textbf{Fidelity and structural agreement across baselines.}}
Constrained contrastive unpaired translation (C-CUT)~\cite{ziabari2025constrained} and the Unpaired Neural Schr{\"o}dinger Bridge (UNSB)~\cite{kim2024unsb} achieve lower FID than \method\ on concrete, but both have lower template MS-SSIM (Table~\ref{tab:latest_mmd}). Main-structure error is 34.49\% for C-CUT and 2.33\% for \method. 
Relative to the unpaired-translation baselines, \method\ substantially reduces hallucination in paste and main structure, and its pore error is lower than C-CUT's and comparable to UNSB's. The pore advantage over C-CUT widens at higher resolution (Table~\ref{tab:higher_resolution}). Methods with lower pore error either stay close to the simulated input or give up fidelity.
Projected Diffusion Models (PDM)~\cite{christopher2024pdm} achieve higher template MS-SSIM than \method\ with spectral guidance, and Physics-Constrained Flow Matching (PCFM)~\cite{utkarsh2026pcfm} with an inferred endpoint nearly matches \method. Both have higher FID (58.108 and 44.216, respectively, versus 14.588). Our Diffusion Translation baseline draws on Zhang et al.~\cite{zhang2023inst}. For each modality, we train a single denoiser on simulated images with a material mask and a synthetic-domain label and on experimental images with only an experimental-domain label. At inference, we add sampled Gaussian noise to the simulated image \(x_0\) and denoise it under the experimental-domain label.

The single-constraint results expose the tradeoff addressed by joint guidance (Fig.~\ref{fig:spectral_effect}). MMD-only sampling gives a FID close to the unconditional sampling; spectral-only sampling raises MS-SSIM to 0.8699 but FID to 71.578. Joint guidance retains MS-SSIM 0.8565 with FID 14.588, yielding $H_R=0.7842$. The spectral constraint supplies per-image layout information, while MMD limits cohort feature discrepancy.

Several \method\ settings improve both metrics over evaluated C-CUT checkpoints (Fig.~\ref{fig:fid_msssim_pareto}). We selected settings on the evaluation cohort with unequal search budgets. The flow comparison uses one seed (Secs.~\ref{sec:experimental_details}, \ref{sec:configuration_selection}, and~\ref{sec:seed_variability}).

\noindent\textbf{Hard and soft boundary constraints.}
We test hard and soft correction with C-CUT and \method. Hard correction copies simulated-image pixels within a label-derived boundary band, preserving them exactly but potentially retaining synthetic texture. Soft correction penalizes differences between predicted segmentation edges and label boundaries through $\lambda_{\rm boundary}$, allowing texture and intensity changes. \method\ predicts these edges with a differentiable Gaussian intensity classifier. Its boundary variants use $\epsilon_{\rm final}=0.02$, versus 0.01 for the standard configuration; Sec.~\ref{sec:boundary_protocol} gives the settings and matched control.
Compared with the standard \method, soft boundary guidance raises MS-SSIM from 0.8565 to 0.9379, leading to the gain of $H_R$ from 0.7842 to 0.8334 even as FID worsens (14.588 vs. 19.275). The variant reduces the paste and main-structure errors, but almost doubles the pore error from 6.10\% to 11.33\%. Hard guidance preserves boundary-band pixels but gives lower template agreement and higher FID.

\noindent\textbf{Higher-resolution XCT microstructure.}
At $512\times512$, \method\ again trades higher FID for stronger template agreement than C-CUT (35.430 vs.\ 26.309 and 0.8077 vs.\ 0.7360; Table~\ref{tab:higher_resolution}). At $1024\times1024$, \method\ reaches FID 89.835 and MS-SSIM 0.7476. At both resolutions, soft boundary guidance reduces paste and main-structure hallucination but increases pore hallucination relative to standard \method. Thus stronger template agreement need not improve pore labels, a distinction relevant to porosity and defect measurements. Figure~\ref{fig:highres_examples} shows both resolutions; Sec.~\ref{sec:highres_protocol} details the cohorts.

\subsection{Ablation Studies}
\label{sec:ablation_studies}
The single-constraint sweeps confirm that tighter spectral budgets strengthen template agreement at an appearance cost: MS-SSIM and FID both increase (Table~\ref{tab:constraint_ablations}). The MMD threshold's effect on FID depends on batch size and is not always monotonic, so tighter cohort constraints alone need not improve this separate fidelity metric.

When \method\ corrects throughout sampling, terminal projection changes FID only from 13.989 to 13.993. Ending intermediate correction early increases the terminal solve's fidelity cost, indicating that continued guidance reduces reliance on a costly terminal correction. Table~\ref{tab:ablations} also examines batch size, budgets, and target smoothing.

\subsection{Comparison Across Imaging Regimes} 
\label{sec:additional_datasets}
We compare \method, C-CUT, UNSB, Diffusion Translation, PDM, and PCFM across synchrotron concrete, laboratory microCT powder, and industrial XCT metal AM using 500 maps per dataset at $256\times256$ (Table~\ref{tab:additional_modalities}). We use the concrete settings from Table~\ref{tab:latest_mmd}; Sec.~\ref{sec:additional_dataset_protocol} describes metal and powder.

\method\ achieves higher template MS-SSIM and $H_R$ than every C-CUT variant across all three regimes, extending the layout advantage across different source and detector physics (Fig.~\ref{fig:additional_modalities}). C-CUT retains lower FID: standard C-CUT is best among its variants on concrete and powder, and soft-boundary C-CUT on metal. On powder, \method\ raises MS-SSIM from 0.8913 to 0.9174 with FID increasing from 9.770 to 10.807.

\subsection{SAM Segmentation}
\label{sec:sam_segmentation}
We fine-tuned the SAM2.1 Large mask decoder~\cite{kirillov2023sam,ravi2025sam2} on 400 generated or simulated XCT images with identical labels and training settings (Sec.~\ref{sec:sam_protocol}). On 94 generated images from held-out slices of the same volume, identical prompts gave the highest mean IoU and Dice after adaptation on generated pairs (0.7506 and 0.8435), followed by simulated pairs (0.7060 and 0.7983) and pretrained SAM2.1 (0.6284 and 0.7295). This supports their use for adaptation within the generated-image domain; transfer to independent real specimens remains untested.

\section{Discussion and Limitations}
\label{sec:discussion}

\noindent\textbf{Terminal projection.}
When intermediate guidance stopped at $t=0.80$, the terminal projection raised FID (Table~\ref{tab:ablations}). Every corrected batch passed its MMD check with earlier uncorrected outputs in the cache, yet the saved cohort exceeded the bound. The paired study did not test online cache updates using the saved images (Sec.~\ref{sec:failure_modes}).

\noindent\textbf{Content hallucination and the fidelity constraint.}
In C-CUT, a change in phase content can improve the adversarial match to the target collection. In synchrotron concrete, C-CUT achieves a lower FID than \method, but a higher fraction of the prescribed main-structure misclassified as pore or paste (Table~\ref{tab:latest_mmd}). \method\ checks the spectral bound in low-frequency agreement with each map and the MMD bound in the discrepancy in the features of the cohort separately. The joint solver does not give bound priority, and finite optimization can also leave a residual. The separate checks let us report which requirement was missed.

\noindent\textbf{Structural agreement and segmentation.}
The spectral bound constrains agreement over the selected low frequencies. A small pore or thin paste ligament can shift even when the image meets the bound (Sec.~\ref{sec:spectral}). The phase hallucination rates measure map disagreement through an intensity classifier. Texture changes can alter its assignments even when a phase boundary stays in place. The SAM study evaluated segmentation on held-out generated images (Sec.~\ref{sec:sam_segmentation}). Whether adaptation on generated pairs improves segmentation on independent real specimens remains open.
\section{Conclusion}
We presented \method, which generates labeled scientific images from truly unpaired material maps and experimental scans by sampling a frozen flow-matching prior under a per-image spectral constraint and a cohort-level MMD constraint. Unlike unpaired translation, it does not require the simulated and real collections to share feature sizes or densities, because content comes from the map and only appearance comes from the data. Across synchrotron, laboratory microCT, and industrial X-ray tomography, the same procedure, without instrument-specific modeling, achieved higher structural agreement than C-CUT at a modest fidelity cost. Because segmentation here is a measurement, label-faithful synthetic data matters beyond benchmark accuracy: it determines the accuracy of downstream porosity, phase fractions, and defect statistics.

\section*{LLM Usage Statement}
We used large language models to assist in drafting and revising the manuscript for clarity and readability. We take full responsibility for reviewing all AI-assisted material and for the final content of the paper.

{
    \small
    \bibliographystyle{ieeenat_fullname}
    \bibliography{main}
}


\clearpage
\appendix
\setcounter{table}{0}
\renewcommand{\thetable}{A\arabic{table}}
\renewcommand{\theHtable}{appendix.\arabic{table}}
\section{Experimental Protocol}
\label{sec:experimental_details}
\subsection{Synchrotron Concrete}
\paragraph{Training and sampling.}
The U-Net has base width 64, channel multipliers $(1,2,4,4)$, and two residual blocks per level. We train with AdamW at batch size 16. The learning rate rises to $2\times10^{-4}$ over 1,000 steps and decays to $2\times10^{-5}$ on a cosine schedule.

The $256\times256$ ablations use the 140k EMA checkpoint, seed 123, and 80 Heun steps. Unless varied, $B=8$, $\gamma=1.25$, and the spectral budget decreases from 0.08 to 0.02 with $c=0.06$ and $w=0.01$. We correct every second step over $t=0.30$ to $1.00$ and project the integrated endpoint before encoding. Each solve allows 50 iterations with primal and dual learning rates of 0.01 and 0.2 and $\sigma_d=0.1$.

\paragraph{Targets and evaluation.}
The calibrated mean intensities on $[0,1]$ are 0.4465 for pore, 0.5917 for paste, and 0.6534 for main structure. Projection and spectral evaluation use a template smoothed with a one-pixel Gaussian. MS-SSIM~\cite{wang2003msssim} uses the unsmoothed template. The MMD reference set contains 2,048 real images disjoint from the pseudo-label source images, and the generated feature cache holds up to 500 images. FID~\cite{heusel2017} and MS-SSIM use 8-bit images. FID uses original Inception-v3 features and float64 covariance statistics unless noted. We use $R=172.862$ throughout and compare $H_R$ within each dataset and resolution.

\subsection{Selection and Baselines}
\label{sec:configuration_selection}
We selected configurations on the same 500-map cohort used for Table~\ref{tab:latest_mmd}. Among the 18 terminal \method\ configurations in the original comparison, $B=8$, $\epsilon_{\rm final}=0.01$, and $\gamma=1.25$ gave the highest $H_R$. Both single-constraint rows use $B=8$: spectral-only retains $\epsilon_{\rm final}=0.01$, and MMD-only retains $\gamma=1.25$. Table~\ref{tab:constraint_ablations} reports the separate sweeps over batch size and constraint budget.

The highest $H_R$ selected epochs 60 and 120 for standard and soft-boundary C-CUT~\cite{ziabari2025constrained}, respectively. All hard-boundary C-CUT checkpoints had $H_R=0$, so we chose epoch 50 by lowest FID. UNSB uses the 200k checkpoint. Figure~\ref{fig:spectral_effect} uses the Table~\ref{tab:latest_mmd} settings and a common grayscale range of $[0,255]$.

PDM~\cite{christopher2024pdm} uses a separate 200k EMA score model and 500 spectrally projected Langevin updates with an unsmoothed template and transition width 0.02. PCFM~\cite{utkarsh2026pcfm} uses 80 Euler steps with our flow prior, spectral projector, target, and schedule. Each correction infers the noise endpoint as $x_t-tv_\theta(x_t,t)$.

Diffusion Translation trains one denoiser per modality on unpaired XCT images. Each image carries a domain label, and only synthetic images have material masks. Zhang et al.~\cite{zhang2023inst} estimate noise from the content image for style transfer. We corrupt the simulated reconstruction $x_0$ from the forward model by injecting noise. We denoise from $x_t$ under the experimental-domain label, starting at $t=300$ for concrete, $200$ for metal, and $400$ for powder.

\paragraph{Boundary variants.}
\label{sec:boundary_protocol}
At $\epsilon_{\rm final}=0.02$, \method\ without boundary correction gives FID 13.910, MS-SSIM 0.8318, and $H_R=0.7649$. At this same budget, soft correction ($\lambda_{\rm boundary}=1$) improves template MS-SSIM to 0.9379 and $H_R$ to 0.8334, while FID worsens to 19.275. Hard correction copies a one-pixel-radius band without blending. All protected pixels matched the 16-bit source.

\subsection{Higher Resolution and Additional Datasets}
\label{sec:highres_protocol}
For Table~\ref{tab:higher_resolution}, \method\ uses the 95k raw checkpoint at $512\times512$ and the 80k EMA checkpoint at $1024\times1024$. C-CUT uses epochs 260 and 400, respectively. One $1024\times1024$ C-CUT file was empty, leaving 499 readable image--map pairs. At that resolution ,we trained the flow-matching prior on shifted crops while C-CUT on unshifted ones. To match the distribution, we computed FID using 3,696 shifted real crops for SF-Flow and 2,816 unshifted real crops for C-CUT.

\begin{figure*}[!tp]
\centering
{\small (a) $512\times512$}\par
\includegraphics[width=\linewidth]{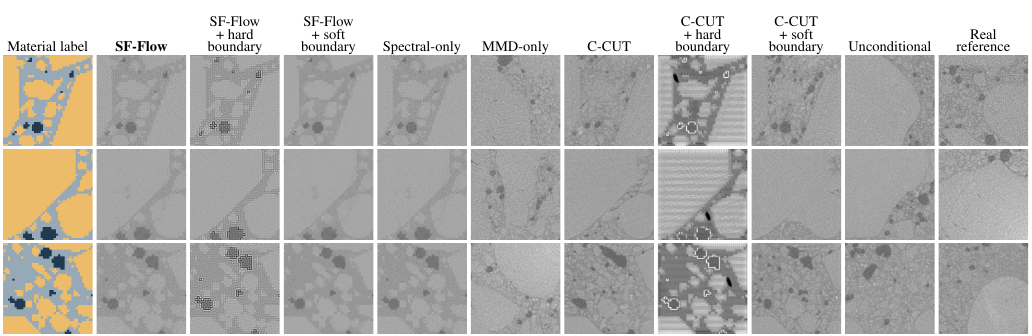}\par
{\small (b) $1024\times1024$}\par
\includegraphics[width=\linewidth]{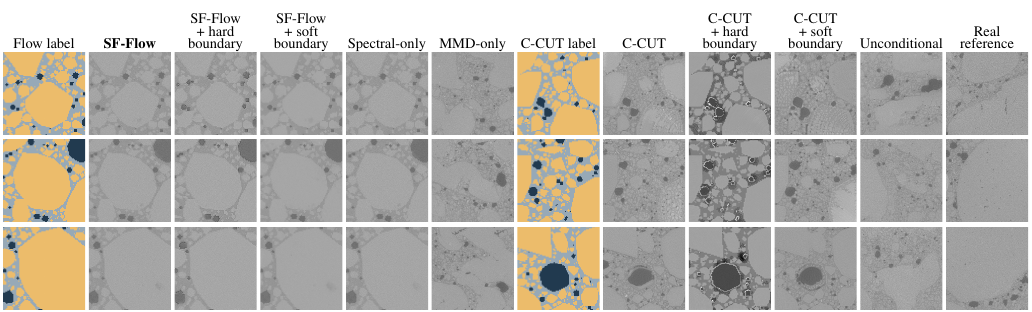}
\caption{Higher-resolution synchrotron concrete synthesis for the methods in Table~\ref{tab:higher_resolution}. Each panel shows three examples, comparing \method, its single-constraint and boundary variants, and C-CUT with its boundary variants. (a) At $512\times512$, methods share the prescribed material map in each row. (b) At $1024\times1024$, \method\ and C-CUT use different crop sets; their corresponding maps are shown in the ``Flow label'' and ``C-CUT label'' columns. The final two columns show unconditional samples and unpaired real references for comparison of texture and imaging appearance, without correspondence to the prescribed layouts.}
\label{fig:highres_examples}
\end{figure*}

\begin{figure*}[!tp]
\centering
\includegraphics[width=\linewidth]{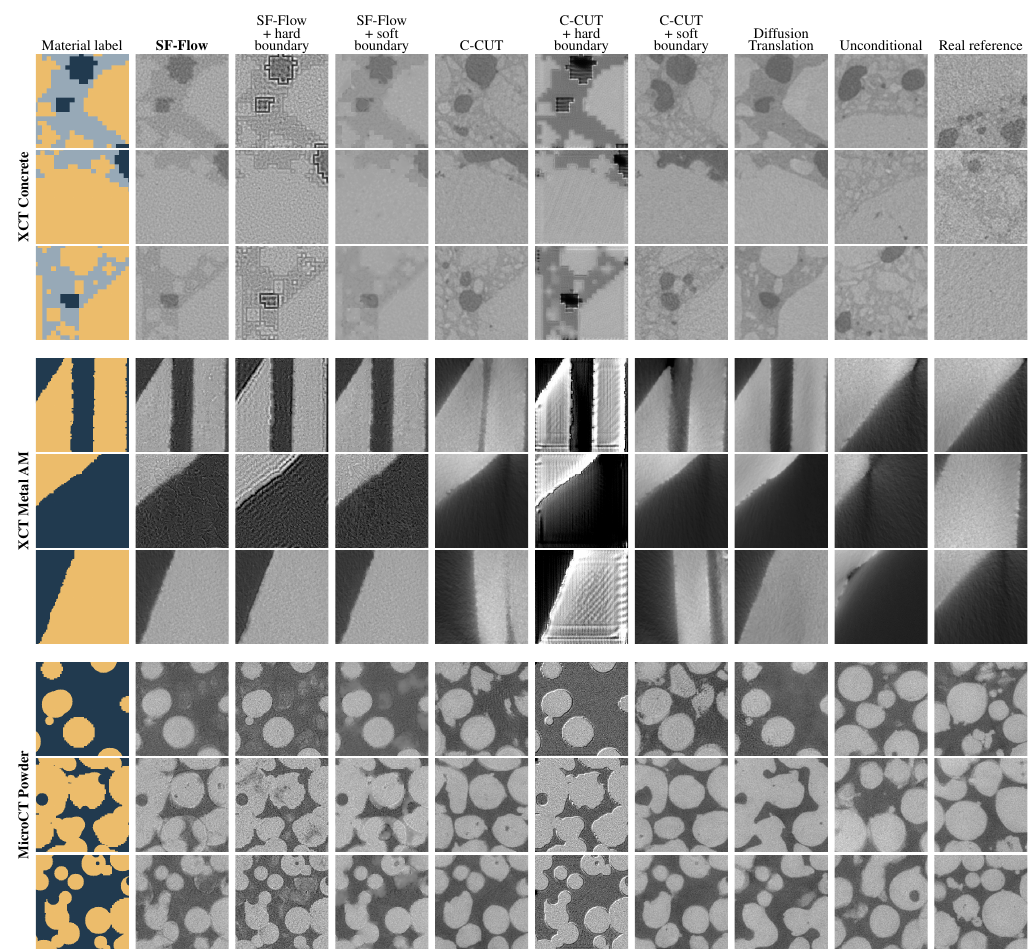}
\caption{Qualitative comparison across the imaging regimes in Table~\ref{tab:additional_modalities}: synchrotron concrete (top), industrial XCT metal AM (middle), and laboratory microCT powder (bottom). Each group shows three prescribed $256\times256$ material maps with outputs from \method, C-CUT, their hard- and soft-boundary variants, and Diffusion Translation. Comparing each output with the leftmost map reveals changes in phase placement and boundaries, while the final two columns provide unconditional samples and unpaired real references for assessing appearance. A common grayscale window is used within each dataset.}
\label{fig:additional_modalities}
\end{figure*}

\paragraph{Metal and powder.}
\label{sec:additional_dataset_protocol}
Each dataset contributes 500 binary maps at $256\times256$. We estimate class means from training image--mask pairs, excluding images paired with test maps. \method\ and PCFM use the 90k flow EMA checkpoint for metal and the 55k flow EMA checkpoint for powder. PDM uses the 200k diffusion EMA checkpoint for each dataset. Both \method\ runs use $\gamma=0.10$, $\epsilon_{\rm final}=0.02$, and 2,048 real MMD references. FID uses 10,000 real metal images and 6,108 real powder images. All C-CUT variants use epoch 400. Fig.~\ref{fig:additional_modalities} uses a common grayscale window within each dataset.

\subsection{SAM Adaptation}
\label{sec:sam_protocol}
We trained the SAM2.1 Large mask decoder for 20 epochs at learning rate $10^{-5}$ and effective batch size eight, keeping the other weights fixed. Within-domain validation selected epoch 14 for generated training images and epoch 20 for simulated training images. All models used CPU float32 for inference and for computing macro IoU and Dice against the original component masks.

\section{Sampling Ablations}
\label{sec:ablations}
\subsection{Constraint Budgets and Seed Variability}
\label{sec:single_constraint_sweeps}
In Table~\ref{tab:constraint_ablations}, tightening the spectral budget raises template MS-SSIM and FID at every batch size. Tightening the MMD threshold consistently lowers FID at $B=1$ and $B=2$ over the range shown. Each batch size uses its own noise draws and MMD calibration.

\paragraph{Fixed-setting repeats.}
\label{sec:seed_variability}
Across ten seeds with 500 maps each ($B=8$, $\epsilon_{\rm final}=0.02$, $\gamma=1.25$ where applicable), mean FID and sample standard deviation were $9.365\pm0.290$ for unconditional sampling, $46.814\pm0.842$ for spectral-only, $9.334\pm0.176$ for MMD-only, and $14.148\pm0.290$ for \method. For Table~\ref{tab:latest_mmd}, we selected \method\ by $H_R$ and matched the single-constraint ablations to its batch size and remaining constraint setting.

\subsection{Joint Sampling and Terminal Projection}
\label{sec:sf_sampling_settings}
Table~\ref{tab:ablations} varies the batch size, budgets, template smoothing width $\sigma$ (pixels), and end of intermediate guidance $t_{\rm end}$. At $B=4$, tightening the MMD threshold lowers both FID and template MS-SSIM at the two tested spectral budgets of 0.02 and 0.03. The best follow-up $H_R$ was 0.7882 ($B=4$, $\epsilon_{\rm final}=0.01$, $\gamma=1$), above the 0.7842 selected in the original comparison.

\begin{table*}[p]
\centering
\captionsetup{skip=4pt}
\caption{Single-constraint sweeps on 500 synchrotron concrete maps at $256\times256$. (a) MMD-only varies the threshold multiplier $\gamma$; (b) spectral-only varies the final residual budget $\epsilon_{\rm final}$. Smaller values tighten the corresponding constraint, and $B$ is the sampling batch size. FID measures cohort fidelity, MS-SSIM measures agreement with the unsmoothed template, and $H_R$ combines both; arrows indicate the preferred direction. Phase hallucination reports the percentage of prescribed pixels assigned another phase. Tighter spectral budgets improve template agreement but raise FID, while the effect of the MMD threshold on FID depends on batch size.}
\label{tab:constraint_ablations}
\footnotesize
\setlength{\aboverulesep}{0.8pt}
\setlength{\belowrulesep}{1.1pt}
\setlength{\tabcolsep}{1pt}
\renewcommand{\arraystretch}{0.82}
\begin{minipage}[t]{0.493\textwidth}

\centering
\textit{(a) MMD-only: threshold factor and batch size}\par\smallskip
\begin{tabular*}{\linewidth}{@{\extracolsep{\fill}}rrrrrrrr@{}}
\toprule
\multicolumn{5}{c}{} & \multicolumn{3}{c}{Phase hallucination (\%, $\downarrow$)}\\
$B$ & $\gamma$ & $H_R\uparrow$ & FID$\downarrow$ & MS-SSIM$\uparrow$ & Pore & Paste & Main struct.\\
\midrule
1 & 0.50 & 0.5517 & 9.150 & 0.5825 & 97.21 & 60.59 & 42.59\\
1 & 0.75 & 0.5520 & 9.176 & 0.5830 & 97.21 & 60.61 & 42.57\\
1 & 1.00 & 0.5514 & 9.354 & 0.5830 & 97.21 & 60.62 & 42.57\\
1 & 1.25 & 0.5509 & 9.510 & 0.5830 & 97.21 & 60.61 & 42.57\\
\midrule
2 & 0.50 & 0.5471 & 8.903 & 0.5768 & 96.63 & 58.93 & 44.72\\
2 & 0.75 & 0.5465 & 9.262 & 0.5774 & 96.62 & 58.94 & 44.70\\
2 & 1.00 & 0.5454 & 9.667 & 0.5777 & 96.62 & 58.94 & 44.69\\
2 & 1.25 & 0.5451 & 9.784 & 0.5778 & 96.62 & 58.95 & 44.68\\
\midrule
4 & 0.50 & 0.5457 & 8.804 & 0.5750 & 94.55 & 59.78 & 44.93\\
4 & 0.75 & 0.5446 & 8.969 & 0.5744 & 95.05 & 59.15 & 44.82\\
4 & 1.00 & 0.5448 & 9.253 & 0.5756 & 95.05 & 59.16 & 44.78\\
4 & 1.25 & 0.5455 & 9.521 & 0.5773 & 95.08 & 59.53 & 44.79\\
\midrule
8 & 0.50 & 0.5269 & 8.992 & 0.5558 & 95.40 & 58.14 & 45.68\\
8 & 0.75 & 0.5397 & 9.411 & 0.5708 & 96.18 & 59.34 & 45.32\\
8 & 1.00 & 0.5439 & 9.431 & 0.5753 & 96.07 & 58.97 & 44.06\\
8 & 1.25 & 0.5476 & 9.300 & 0.5788 & 95.05 & 60.24 & 44.72\\
\midrule
16 & 0.50 & 0.5088 & 9.714 & 0.5391 & 96.54 & 57.67 & 45.74\\
16 & 0.75 & 0.5260 & 9.318 & 0.5560 & 96.56 & 58.05 & 45.41\\
16 & 1.00 & 0.5294 & 9.530 & 0.5603 & 96.57 & 58.07 & 45.38\\
16 & 1.25 & 0.5321 & 9.939 & 0.5646 & 96.60 & 58.10 & 45.28\\
\bottomrule
\end{tabular*}
\end{minipage}\hfill
\begin{minipage}[t]{0.493\textwidth}

\centering
\textit{(b) Spectral-only: residual budget and batch size}\par\smallskip
\begin{tabular*}{\linewidth}{@{\extracolsep{\fill}}rrrrrrrr@{}}
\toprule
\multicolumn{5}{c}{} & \multicolumn{3}{c}{Phase hallucination (\%, $\downarrow$)}\\
$B$ & $\epsilon_{\rm final}$ & $H_R\uparrow$ & FID$\downarrow$ & MS-SSIM$\uparrow$ & Pore & Paste & Main struct.\\
\midrule
1 & 0.01 & 0.4931 & 73.193 & 0.8551 & 5.63 & 5.29 & 2.27\\
1 & 0.02 & 0.5738 & 53.901 & 0.8338 & 7.37 & 7.89 & 3.49\\
1 & 0.03 & 0.6213 & 40.027 & 0.8085 & 10.19 & 12.87 & 6.72\\
1 & 0.05 & 0.6626 & 21.375 & 0.7561 & 32.21 & 29.52 & 18.42\\
\midrule
2 & 0.01 & 0.4920 & 74.017 & 0.8603 & 5.72 & 4.91 & 2.22\\
2 & 0.02 & 0.5898 & 51.462 & 0.8399 & 7.95 & 7.29 & 3.42\\
2 & 0.03 & 0.6398 & 36.989 & 0.8140 & 11.45 & 11.95 & 6.45\\
2 & 0.05 & 0.6666 & 21.045 & 0.7590 & 35.06 & 27.86 & 17.75\\
\midrule
4 & 0.01 & 0.5037 & 72.210 & 0.8651 & 5.89 & 4.67 & 2.18\\
4 & 0.02 & 0.6042 & 48.999 & 0.8433 & 8.41 & 6.90 & 3.43\\
4 & 0.03 & 0.6530 & 34.777 & 0.8174 & 12.71 & 11.77 & 6.25\\
4 & 0.05 & 0.6555 & 22.985 & 0.7560 & 33.02 & 27.48 & 18.15\\
\midrule
8 & 0.01 & 0.5097 & 71.578 & 0.8699 & 6.21 & 4.49 & 2.20\\
8 & 0.02 & 0.6119 & 47.839 & 0.8461 & 9.09 & 6.84 & 3.44\\
8 & 0.03 & 0.6615 & 33.165 & 0.8185 & 14.28 & 11.64 & 6.28\\
8 & 0.05 & 0.6537 & 22.646 & 0.7523 & 29.31 & 28.41 & 17.34\\
\midrule
16 & 0.01 & 0.5176 & 70.611 & 0.8751 & 6.78 & 4.41 & 2.22\\
16 & 0.02 & 0.6250 & 46.100 & 0.8523 & 10.77 & 6.75 & 3.52\\
16 & 0.03 & 0.6723 & 31.789 & 0.8238 & 16.96 & 11.36 & 6.08\\
16 & 0.05 & 0.6593 & 21.914 & 0.7551 & 34.70 & 27.36 & 16.11\\
\bottomrule
\end{tabular*}
\end{minipage}
\par\smallskip

\caption{\method\ sampling ablations on 500 synchrotron concrete maps at $256\times256$. Columns specify batch size $B$, final spectral budget $\epsilon_{\rm final}$, MMD threshold multiplier $\gamma$, template smoothing width $\sigma$ in pixels, and the final time $t_{\rm end}$ of intermediate guidance. ``Terminal'' indicates a joint projection at $t=1$ before image encoding. Metrics assess cohort fidelity (FID), template agreement (MS-SSIM), their combined score ($H_R$), and phase misclassification percentages. Ending intermediate guidance early substantially increases the FID cost of terminal projection. Bold identifies the configuration used in Table~\ref{tab:latest_mmd}, selected before the tighter-budget follow-ups.}
\label{tab:ablations}
\begin{tabular*}{\textwidth}{@{\extracolsep{\fill}}rrrrrcrrrrrr@{}}
\toprule
\multicolumn{9}{c}{} & \multicolumn{3}{c}{Phase hallucination (\%, $\downarrow$)}\\
$B$ & $\epsilon_{\rm final}$ & $\gamma$ & $\sigma$ & $t_{\rm end}$ & Terminal & $H_R\uparrow$ & FID$\downarrow$ & MS-SSIM$\uparrow$ & Pore & Paste & Main struct.\\
\midrule
\multicolumn{12}{@{}l}{\textit{Terminal correction and intermediate projection end time}}\\
8 & 0.02 & 1.25 & 1 & 1.00 & No & 0.7657 & 13.989 & 0.8331 & 8.95 & 7.03 & 3.67\\
8 & 0.02 & 1.25 & 1 & 1.00 & Yes & 0.7657 & 13.993 & 0.8331 & 8.95 & 7.03 & 3.66\\
8 & 0.02 & 1.25 & 1 & 0.80 & No & 0.7579 & 12.999 & 0.8195 & 7.64 & 7.95 & 4.18\\
8 & 0.02 & 1.25 & 1 & 0.80 & Yes & 0.3653 & 96.856 & 0.8309 & 6.90 & 6.70 & 3.27\\
8 & 0.02 & 1.25 & 1 & 0.65 & No & 0.7069 & 14.268 & 0.7705 & 7.04 & 10.18 & 5.61\\
8 & 0.02 & 1.25 & 1 & 0.65 & Yes & 0.0000 & 240.046 & 0.7997 & 6.04 & 7.26 & 3.36\\
\midrule
\multicolumn{12}{@{}l}{\textit{Sampling batch size}}\\
1 & 0.02 & 1.25 & 1 & 1.00 & Yes & 0.7501 & 17.035 & 0.8321 & 7.26 & 7.85 & 3.48\\
2 & 0.02 & 1.25 & 1 & 1.00 & Yes & 0.7639 & 15.056 & 0.8368 & 7.82 & 7.33 & 3.45\\
4 & 0.02 & 1.25 & 1 & 1.00 & Yes & 0.7696 & 14.125 & 0.8381 & 8.43 & 6.96 & 3.52\\
16 & 0.02 & 1.25 & 1 & 1.00 & Yes & 0.7395 & 13.940 & 0.8043 & 9.76 & 7.53 & 4.31\\
\midrule
\multicolumn{12}{@{}l}{\textit{Final spectral budget}}\\
\textbf{8} & \textbf{0.01} & \textbf{1.25} & \textbf{1} & \textbf{1.00} & \textbf{Yes} & \textbf{0.7842} & \textbf{14.588} & \textbf{0.8565} & \textbf{6.10} & \textbf{4.75} & \textbf{2.33}\\
8 & 0.03 & 1.25 & 1 & 1.00 & Yes & 0.7398 & 13.388 & 0.8019 & 13.19 & 11.61 & 6.46\\
\midrule
\multicolumn{12}{@{}l}{\textit{Target smoothing}}\\
8 & 0.02 & 1.25 & 0 & 1.00 & Yes & 0.7635 & 14.061 & 0.8312 & 8.70 & 6.95 & 3.69\\
8 & 0.02 & 1.25 & 2 & 1.00 & Yes & 0.7599 & 14.384 & 0.8288 & 9.83 & 7.42 & 3.91\\
8 & 0.02 & 1.25 & 3 & 1.00 & Yes & 0.7565 & 14.181 & 0.8241 & 11.49 & 7.82 & 4.28\\
\midrule
\multicolumn{12}{@{}l}{\textit{MMD threshold at two spectral budgets}}\\
4 & 0.02 & 0.25 & 1 & 1.00 & Yes & 0.7567 & 11.324 & 0.8097 & 8.33 & 7.66 & 4.01\\
4 & 0.02 & 0.5 & 1 & 1.00 & Yes & 0.7673 & 12.374 & 0.8264 & 8.36 & 7.24 & 3.65\\
4 & 0.02 & 1 & 1 & 1.00 & Yes & 0.7712 & 13.623 & 0.8372 & 8.46 & 7.05 & 3.46\\
4 & 0.03 & 0.25 & 1 & 1.00 & Yes & 0.7335 & 10.929 & 0.7830 & 12.52 & 12.08 & 6.97\\
4 & 0.03 & 0.5 & 1 & 1.00 & Yes & 0.7441 & 11.848 & 0.7989 & 12.73 & 12.04 & 6.64\\
4 & 0.03 & 1 & 1 & 1.00 & Yes & 0.7472 & 13.241 & 0.8092 & 12.82 & 11.70 & 6.36\\
\midrule
\multicolumn{12}{@{}l}{\textit{Tighter spectral budgets}}\\
4 & 0.01 & 0.5 & 1 & 1.00 & Yes & 0.7879 & 12.737 & 0.8506 & 5.82 & 4.89 & 2.34\\
4 & 0.01 & 1 & 1 & 1.00 & Yes & 0.7882 & 14.290 & 0.8592 & 5.84 & 4.71 & 2.22\\
4 & 0.015 & 0.5 & 1 & 1.00 & Yes & 0.7775 & 12.682 & 0.8391 & 6.93 & 5.77 & 2.87\\
4 & 0.015 & 1 & 1 & 1.00 & Yes & 0.7811 & 13.888 & 0.8493 & 6.90 & 5.57 & 2.70\\
8 & 0.01 & 0.5 & 1 & 1.00 & Yes & 0.7686 & 12.755 & 0.8298 & 5.98 & 5.39 & 2.82\\
8 & 0.01 & 1 & 1 & 1.00 & Yes & 0.7833 & 14.260 & 0.8537 & 6.04 & 4.79 & 2.38\\
8 & 0.015 & 0.5 & 1 & 1.00 & Yes & 0.7601 & 12.578 & 0.8197 & 7.07 & 6.19 & 3.33\\
8 & 0.015 & 1 & 1 & 1.00 & Yes & 0.7736 & 13.841 & 0.8409 & 7.21 & 5.72 & 2.93\\
\bottomrule
\end{tabular*}
\end{table*}

\paragraph{Verification after terminal projection.}
\label{sec:failure_modes}
The paired terminal runs hold the pre-correction feature history fixed. When intermediate guidance continued through $t=1$, terminal correction changed FID from 13.989 to 13.993. When guidance ended at $t=0.80$, the same correction raised FID from 12.999 to 96.856. At $t_{\rm end}=0.65$, one batch exhausted all 50 iterations without a jointly feasible candidate.

\end{document}